\documentclass[twocolumn]{article}
\usepackage[utf8]{inputenc}

\usepackage[a4paper, total={6.5in, 9in}]{geometry}

\usepackage{natbib}
\usepackage[pdfborder={0 0 0}]{hyperref}
\usepackage{todonotes}
\usepackage[loadonly]{enumitem}
\usepackage{multirow}
\usepackage{subcaption}
\usepackage[section]{placeins}
\usepackage{float}

\usepackage{listings}

\newlist{inlinelist}{enumerate*}{1}
\setlist*[inlinelist,1]{label=(\roman*)}

\begin{document}
\title{Development and evaluation of a deep learning model for protein-ligand binding affinity prediction}
\author{%
  Marta M. Stepniewska-Dziubinska\textsuperscript{1}%
  \and Piotr Zielenkiewicz\textsuperscript{1,2}%
  \and Pawel Siedlecki\textsuperscript{1,2,*}%
  }
\date{%
\small
    \textsuperscript{1}Institute of Biochemistry and Biophysics, Polish Academy of Sciences, Pawinskiego 5a, 02-106 Warsaw, Poland\\%
    \textsuperscript{2}Department of Systems Biology, Institute of Experimental Plant Biology and Biotechnology, University of Warsaw, Miecznikowa 1, 02-096 Warsaw, Poland\\[2ex]%
    \textsuperscript{*}pawel@ibb.waw.pl
}

%%%%%%%%%%%%%%%%%%%%%%%%%%%%%%%%%%%%%%%%%%%%%%%%%%%%%%%%%%%%%%%%%%%%%%%%%%%%%
% ABSTRACT
%%%%%%%%%%%%%%%%%%%%%%%%%%%%%%%%%%%%%%%%%%%%%%%%%%%%%%%%%%%%%%%%%%%%%%%%%%%%%

\maketitle

\abstract{
Structure based ligand discovery is one of the most successful approaches for augmenting the drug discovery process.
Currently, there is a notable shift towards machine learning (ML) methodologies to aid such procedures.
Deep learning has recently gained considerable attention as it allows the model to ``learn'' to extract features that are relevant for the task at hand.\\
We have developed a novel deep neural network estimating the binding affinity of ligand-receptor complexes.
The complex is represented with a 3D grid, and the model utilizes a 3D convolution to produce a feature map of this representation, treating the atoms of both proteins and ligands in the same manner.
Our network was tested on the CASF ``scoring power'' benchmark and Astex Diverse Set and outperformed classical scoring functions.\\
The model, together with usage instructions and examples, is available as a git repository at \href{http://gitlab.com/cheminfIBB/pafnucy}{http://gitlab.com/cheminfIBB/pafnucy}\\}

%%%%%%%%%%%%%%%%%%%%%%%%%%%%%%%%%%%%%%%%%%%%%%%%%%%%%%%%%%%%%%%%%%%%%%%%%%%%%
% INTRODUCTION
%%%%%%%%%%%%%%%%%%%%%%%%%%%%%%%%%%%%%%%%%%%%%%%%%%%%%%%%%%%%%%%%%%%%%%%%%%%%%

\section{Introduction}\label{sec:intro}
Structure-based virtual screening techniques are some of the most successful methods for augmenting the drug discovery process \citep{vsFradera2017,vsBajusz2017}.
With structure-based screening, one tries to predict binding affinity (or more often, a score related to it) between a target and a candidate molecule based on a 3D structure of their complex.
This allows to rank and prioritize molecules for further processing and subsequent testing.
Numerous scoring schemes have been developed to aid this process, and most of them use statistical and/or expert analysis of available protein-ligand structures \citep{sfVerdonk2003,sfMuegge2006,sfMorris2009}. 
Currently, there is a notable shift towards scoring functions using machine learning (ML) methodologies, and this have been highlighted by several reviews \citep{revCheng2012,revMa2013,revLima2016}.
These methods are naturally capable of capturing non-linear and complex relationships in the available data. 

Rather than ``manually'' creating rules using expert knowledge and statistical inference, ML models use arbitrary functions with adjustable parameters that are capable of transforming the input (in this scenario, a protein-ligand complex) to the output (a score related to protein-ligand binding affinity).
Briefly, when the model is presented with examples of input data paired with the desired outcome, it ``learns'' to return predictions that are in agreement with the values provided.
Typically the process of learning is incremental; by introducing small changes to the model parameters, the prediction is moved closer to the target value.
Prime examples of ML scoring functions are RF-Score \citep{Ballester2010}, which uses random forest, and NNscore \citep{Durrant2010,Durrant2011}, which uses an ensemble of shallow neural networks.
These scoring functions were proven useful in virtual screening campaigns and yielded more active compounds than their classical counterparts \citep{Kinnings2011,Wojcikowski2017}.

However, one drawback of such ML approaches is that they still rely on feature engineering, i.e., they utilize expert knowledge to define rules that will become the basis of input data preprocessing.
Hence, one can argue that they are just more sophisticated classical scoring functions with more complex rules.

The ML rule of thumb says that in order to establish a good predictive model, the model needs a lot of data to be able to distinguish more general trends and patterns from noise.
The growing amount of both structural data and affinity measurements has allowed researchers to explore deep learning.
Briefly, a deep neural network consists of multiple layers of non-linear transformations that extract and combine information from data to develop sophisticated relationships between the input and the output.
One of the main advantages of deep learning is that it allows for the reduction of feature engineering: the model learns to extract features as a natural consequence of the process of fitting the model's parameters to the available data.
It is clear that choosing the representation of the input data has a profound impact on the predictive power of a model.
Currently, there is a lot of effort in the field to incorporate feature extraction directly into the ML model.
In such an approach, a learnable molecule representation replaces classical descriptors and fingerprints and becomes the first part of the model.
Then, this representation is trained together with the predictive part of the model to extract features that are useful in solving a specific task. 
With such a design, it is therefore theoretically possible to find and quantify relationships and/or mechanisms that have not yet been discovered or are unknown to the experts \citep{dlZhang2017,dlNketia2017}.

Deep learning has been relatively widely used by the bioinformatics \citep{seqLeung2014,seqAlipanahi2015,seqPark2015,seqJurtz2017,strJimenez2017} and computational biology community \citep{seqAngermueller2016}.
Several promising examples of deep learning methods have also been shown for computer-aided drug design (CADD) \citep{ligLusci2013,Dahl2014,Ma2015,ligXu2015,Duvenaud2015,Kearnes2016,Jastrzebski2016,Segler2017,Olivecrona2017,GomezBombarelli2017,ligLenselink2017,ligXu2017,ligRamsundar2017,Wallach2015,Ragoza2016,strPereira2016,Gomes2017}.
Although deep learning is more readily used in ligand-based regimes, there are currently a couple of interesting examples of structure-based neural networks.

In AtomNet \citep{Wallach2015}, input -- molecular complex -- is discretized to a 3D grid and fed directly into a convolutional neural network.
Instead of data preprocessing, the model uses a learnable representation to recognize different groups of interacting atoms.
AtomNet is a classification method that yields 1 if the ligand is active and yields 0 otherwise.
Another similar model was created by \citet{Ragoza2016} and trained to perform two independent classification tasks: activity and pose prediction.
However, with classification methods, we lose information about the strength of the interaction between the protein and the ligand.

Since neural networks are also suitable for regression, \citet{Gomes2017} created a model predicting the energy gap between a bounded protein-ligand complex and an unbounded state.
In their work, radial pooling filters with learnable mean and variance were used to process the input.
Such filters enabled the production of a summary of the atom's environment and a representation that was invariant to atom ordering and the orientation of the complex.

Taking into account the current findings and aforementioned approaches, we have developed Pafnucy (pronounced ``paphnusy'') -- a novel deep neural network tailored for many structure-based approaches, including derivative prioritization and virtual screening.
Similar to \citet{Ragoza2016}, the input structure is represented with a 3D grid, and a combination of convolutional and dense layers is used; however, our model tries to predict the exact binding affinity value.
Pafnucy utilizes a more natural approach to atom description in which both proteins and ligands have the same atom types.
This approach serves as a regularization technique as it forces the network to discover general properties of interactions between proteins and ligands.
Additionally, the design of Pafnucy provides insight into the feature importance and information extraction that is done during learning and the final prediction of binding affinity.
The network was implemented with TensorFlow \citep{Abadi2015} using Python API and trained on the PDBbind database \citep{Liu2017}.
The source code, trained model and usage instructions are available as a git repository at \href{http://gitlab.com/cheminfIBB/pafnucy}{http://gitlab.com/cheminfIBB/pafnucy}.
%%%%%%%%%%%%%%%%%%%%%%%%%%%%%%%%%%%%%%%%%%%%%%%%%%%%%%%%%%%%%%%%%%%%%%%%%%%%%
% METHODS
%%%%%%%%%%%%%%%%%%%%%%%%%%%%%%%%%%%%%%%%%%%%%%%%%%%%%%%%%%%%%%%%%%%%%%%%%%%%%

\section{Methods}\label{sec:methods}

\subsection{Data}\label{sec:data}

\subsubsection{Representation of a molecular complex}\label{sec:repr}
Three-dimensional structures of protein-ligand complexes require specific transformations and encoding in order to be utilized by a neural network.
In our approach, we cropped the complex to a defined size of 20-\r{A} cubic box focused at the geometric centre of a ligand.
We then discretized the positions of heavy atoms using a 3D grid with 1-\r{A} resolution (see Supplementary Figure~S1).
This approach allowed for the representation of the input as a 4D tensor in which each point is defined by Cartesian coordinates (the first 3 dimensions of the tensor) and a vector of features (the last dimension).

In Pafnucy, 19 features were used to describe an atom:
\begin{itemize}
  \item 9 bits (one-hot or all null) encoding atom types: \emph{B}, \emph{C}, \emph{N}, \emph{O}, \emph{P}, \emph{S}, \emph{Se}, \emph{halogen}, and \emph{metal} 
  \item 1 integer (1, 2, or 3) with atom hybridization: \emph{hyb}
  \item 1 integer counting the numbers of bonds with other heavyatoms: \emph{heavy\_valence}
  \item 1 integer counting the numbers of bonds with other heteroatoms: \emph{hetero\_valence}
  \item 5 bits (1 if present) encoding properties defined with SMARTS patterns: \emph{hydrophobic}, \emph{aromatic}, \emph{acceptor}, \emph{donor}, and \emph{ring}
  \item 1 float with partial charge: \emph{partialcharge}
  \item 1 integer (1 for ligand, -1 for protein) to distinguish between the two molecules: \emph{moltype}
\end{itemize}

The SMARTS patterns were defined the same way as in our previous project \citep{Stepniewska_Dziubinska2017}.
The partial charges were scaled by the training set's standard deviation in order to get a distribution with a unit standard deviation, which improves learning.
In case of collisions (multiple atoms in a single grid point), which rarely occur for a 1-\r{A} grid, features from all colliding atoms were added.

\subsubsection{Dataset preparation}\label{sec:pdbbind}
The network was trained and tested with protein-ligand complexes from the PDBbind database v. 2016 \citep{Liu2017}.
This database consists of 3D structures of molecular complexes and their corresponding binding affinities expressed with $pK_a$ ($-\log K_d$ or $-\log K_i$) values.
PDBBind complexes were divided by Liu~\emph{et~al.} into 3 overlapping subsets.
The \emph{general set} includes all available data.
From this set, the \emph{refined set}, which comprises complexes with higher quality, is subtracted.
Finally, the complexes from the refined set are clustered by protein similarity, and 5 representative complexes are selected from each cluster.
This fraction of the database is called the \emph{core set} and is designed as a high-quality benchmark for structure-based CADD methods.

To properly employ PDBbind information and prevent data leakage, we have split the data into disjoint subsets, i.e., the refined set was subtracted from the general set, and the core set was subtracted from the refined set so that there are no overlaps between the three subsets.
Next, we have discarded all protein-protein, protein-nucleic acid, and nucleic acid-ligand complexes from these new datasets.
Finally, in order to evaluate our model with the CASF ``scoring power'' benchmark \citep{Li2014}, we needed to exclude all data that overlap with the 195 complexes used in CASF.
We therefore excluded a total of 87 overlapping complexes (5 were part of the general set, and 82 were part of the refined set) from the training and validation sets.

All complexes used in this study were protonated and charged using UCSF Chimera \citep{Pettersen2004} with Amber ff14SB for standard residues and AM1-BCC for non-standard residues and ligands.
No additional improvements nor calibration was performed on the complexes; this default protocol was chosen to be in line with \citep{Li2014} to be able to compare Pafnucy to other methods tested on the CASF ``scoring power'' benchmark.

The remaining complexes of the PDBbind v. 2016 dataset were divided as follows:
\begin{inlinelist}
  \item 1000 randomly selected complexes from the refined set were used in validation,
  \item the whole core set (290 complexes) was used as an external test set,
  \item all other complexes (remainder of the refined set and the general set, 11906 in total) were used as the training set.
\end{inlinelist}
In summary, the general and refined sets were used to train the model and select the hyperparameters, while the core set was used as an external test set that was unknown to the model during training and validation.
The scheme illustrating relationships between the subsets is available in Supplementary Figure~S2.

Atomic features were calculated using Open Babel \citep{OBoyle2011}, and the complexes were transformed into grids.
Helper functions used to prepare the data and Jupyter Notebook with all preprocessing steps are available at \href{http://gitlab.com/cheminfIBB/pafnucy}{http://gitlab.com/cheminfIBB/pafnucy}.

As an additional external test set, we used 73 complexes from the Astex Diverse Set \citep{astexHartshorn2007}.
Of 85 complexes in the original database, we excluded those without binding affinity (11 complexes) and those present in the PDBbind database (a single complex, PDB ID: 1YVF, was present in the general set).
The remaining structures were prepared the same way as the PDBbind database.
This dataset was used in order to test Pafnucy on structures from a different source.

\subsection{Network}\label{sec:net}

\subsubsection{Architecture}\label{sec:arch}
\begin{figure}[H]
  \centering
  \includegraphics[height=0.9\textheight]{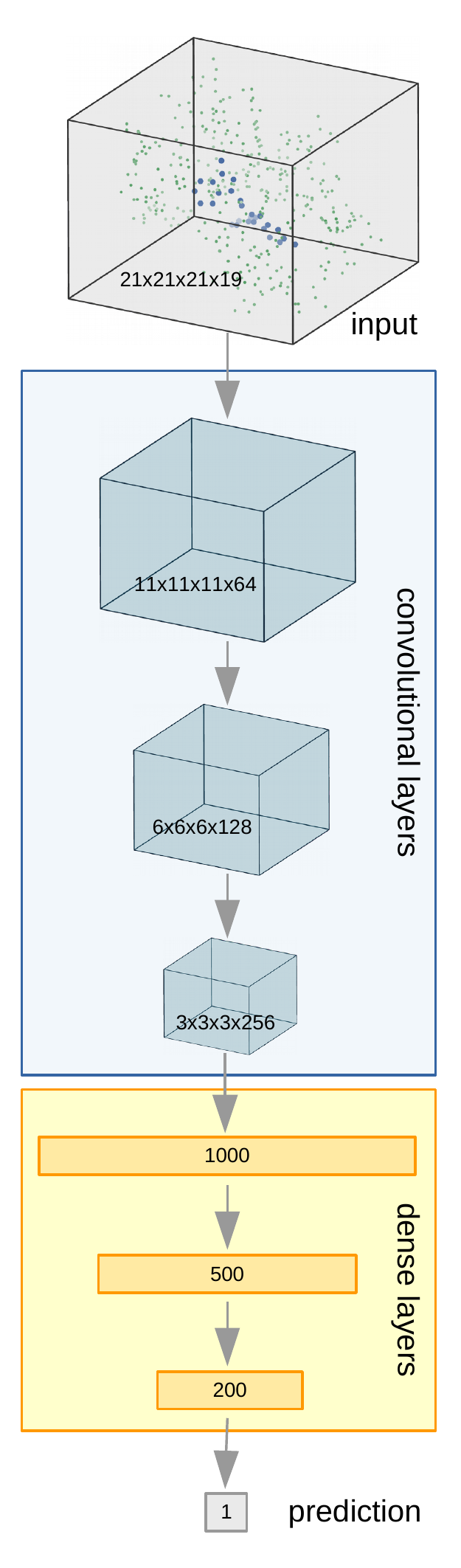}
  \caption[Pafnucy's architecture]{Pafnucy's architecture. The molecular complex is represented with a 4D tensor, processed by 3 convolutional layers and 3 dense (fully-connected) layers to predict the binding affinity.}
  \label{fig:net}
\end{figure}

The architecture used in Pafnucy is a deep convolutional neural network with a single output neuron for predicting the binding affinity.
The model consists of two parts: the convolutional and dense parts, with different types of connections between layers (see Figure~\ref{fig:net}).
Convolution, from which the name ``convolutional'' stems, is a mathematical operation that mixes two functions together.
Most neural network libraries actually substitute the convolution operation with cross-correlation \citep{Goodfellow2016}, which has a more intuitive interpretation and measures the similarity of two functions.
The model discovers patterns that are encoded by the filters in the convolutional layer and creates a feature map with spatial occurrences for each pattern in the data.

Pafnucy's input -- molecular complex -- is represented with a 4D tensor and treated like a 3D image with multiple colour channels.
Each position of an input (x, y, and z coordinates) is described by a vector of 19 properties (see Section~\ref{sec:repr}), which is analogous to how each pixel of an image (x and y coordinates) is described by a vector of intensities of 3 basic colours.

First, the input is processed by a block of 3D convolutional layers combined with a max pooling layer. %SENIOR EDITOR: Please ensure that the above edit maintains the original intended meaning. 
Pafnucy uses 3 convolutional layers with 64, 128, and 256 filters.
Each layer has 5-\r{A} cubic filters and is followed by a max pooling layer with a 2-\r{A} cubic patch.
The result of the last convolutional layer is flattened and used as input for a block of dense (fully-connected) layers.
We used 3 dense layers with 1000, 500, and 200 neurons.
In order to improve generalization, dropout with drop probability of 0.5 was used for all dense layers.
We also experimented with 0.2 dropout and no dropout and achieved worse results on the validation set.

Both convolutional and dense layers are composed of rectified linear units (ReLU).
ReLU was chosen because it speeds up the learning process compared with other types of activations.
We also experimented with Tanh units and achieved a very similar prediction accuracy, but learning was much slower.

\subsubsection{Training}\label{sec:train}
The initial values of the convolutional filter weights were drawn from a truncated normal distribution with 0 mean and 0.001 standard deviation and corresponding biases were set to 0.1.
The weights in the dense layers were initialized with a truncated normal distribution with 0 mean and a standard deviation of $1/\sqrt[]{n}$, where $n$ is the number of incoming neurons for a given layer.
The corresponding biases were set to 1.0.

The Adam optimizer was used to train the network with a $10^{-5}$ learning rate and 5 examples per mini-batch\footnote{The training set contains 11906 complexes; therefore, the last batch actually consisted of 6 complexes instead of 5.}.
Larger batch sizes (10 and 20 examples) were also tested but resulted in worse performance.
Training was carried out for 20 epochs, and the model with the lowest error on the validation set was selected (in the case of the network described in this work, it was after 14 epochs of training).

To reduce overfitting we, used the dropout approach mentioned earlier and L2 weight decay with $\lambda=0.001$.
Using a higher value ($\lambda=0.01$) decreased the model's capacity too much and resulted in higher training and validation errors.
In addition to providing regularization, L2 allows us to investigate feature importance.
If a weight differs from 0 considerably, the information it transfers must be important for the model to make a prediction (see Section~\ref{sec:discussion}).

An important part of our approach was to develop a model that was not sensitive to ligand-receptor complex orientation.
Therefore every structure was presented to the network in 24 different orientations (i.e., all possible combinations of $90^{\circ}$ rotations of a cubic box), yielding 24 different training examples per protein-ligand complex.

By using systematic rotations of complexes during training, we anticipated that the network would learn more general rules about protein-ligand interactions and lead to better performance on new data.
Indeed, in our experiments, we observed a much worse performance of models trained on single orientations regardless of the hyperparameters used to define a particular network.

%%%%%%%%%%%%%%%%%%%%%%%%%%%%%%%%%%%%%%%%%%%%%%%%%%%%%%%%%%%%%%%%%%%%%%%%%%%%%
% RESULTS
%%%%%%%%%%%%%%%%%%%%%%%%%%%%%%%%%%%%%%%%%%%%%%%%%%%%%%%%%%%%%%%%%%%%%%%%%%%%%

\section{Results}\label{sec:results}
The error on training and validation sets was monitored during learning (see Supplementary Figure~S3).
Although the model was trained on 24 different rotations of each complex, the $RMSE$ (root mean square error) was calculated for the original orientation only in order to speed up the computations.

\begin{figure*}[!htb]
  \centering
  \begin{subfigure}[b]{0.45\textwidth}
    \includegraphics[width=\textwidth]{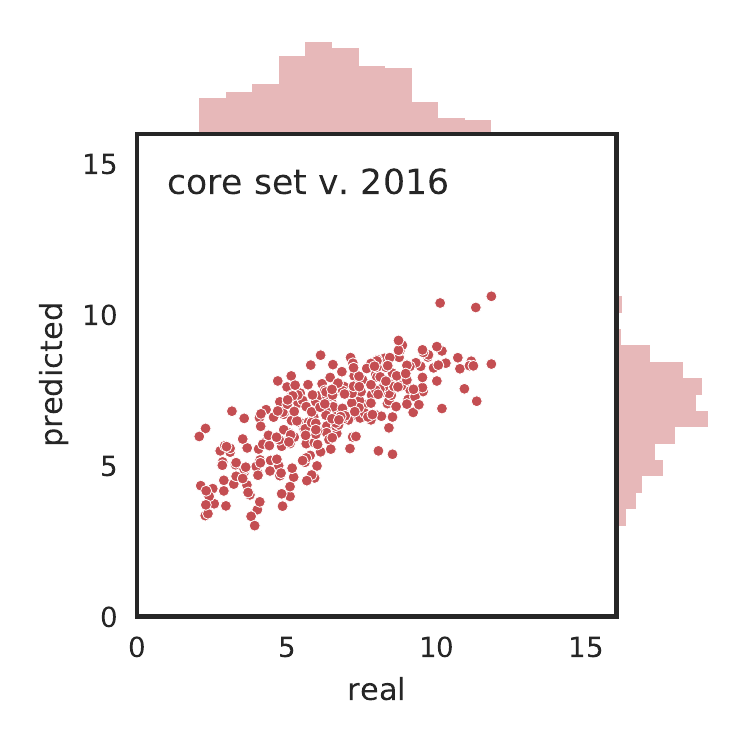}
    \caption{v. 2016 core set; $RMSE=1.42$, $R=0.78$}
    \label{fig:pred-test}
  \end{subfigure}
~
  \begin{subfigure}[b]{0.45\textwidth}
    \includegraphics[width=\textwidth]{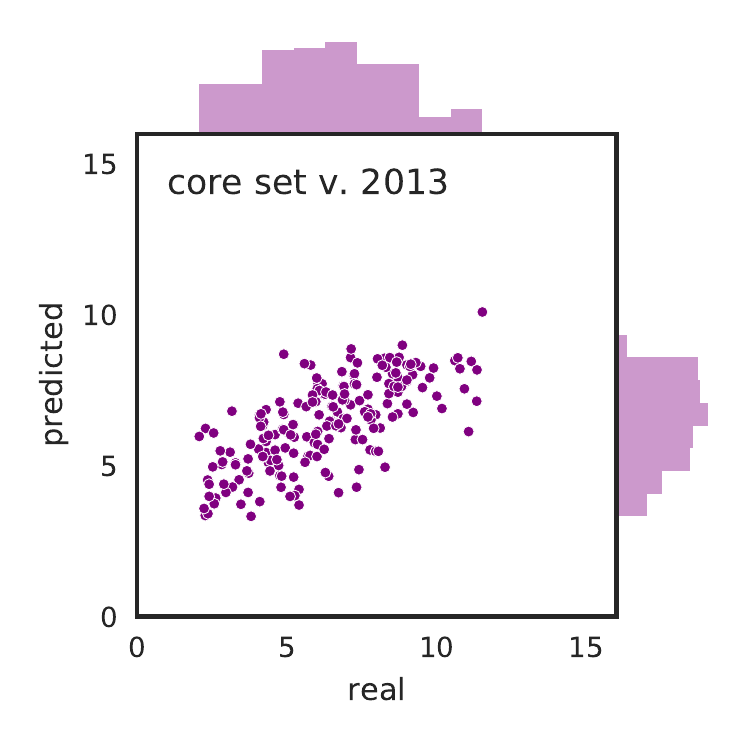}
    \caption{v. 2013 core set; $RMSE=1.62$, $R=0.70$}
    \label{fig:pred-test2013}
  \end{subfigure}
  
  \begin{subfigure}[b]{0.45\textwidth}
    \includegraphics[width=\textwidth]{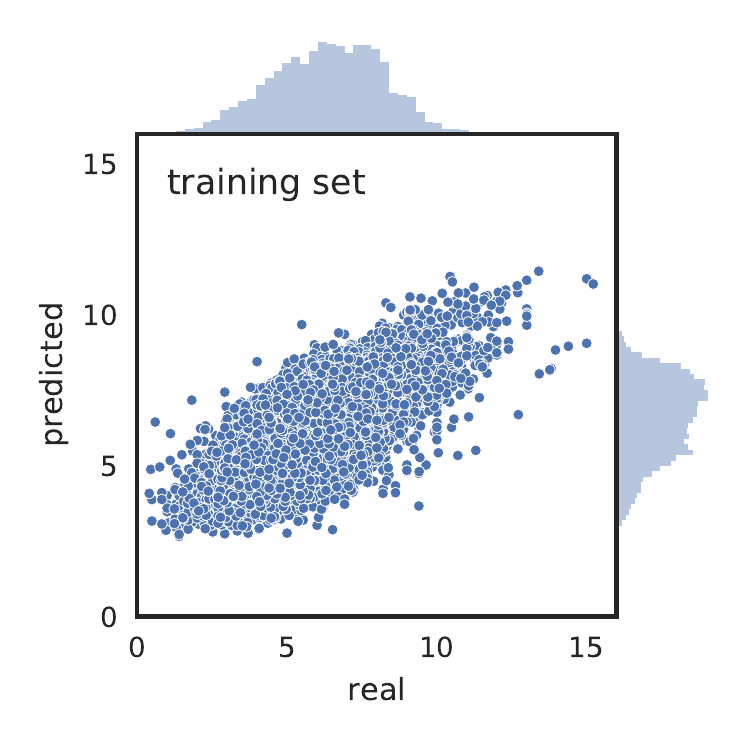}
    \caption{training set; $RMSE=1.21$, $R=0.77$}
    \label{fig:pred-train}
  \end{subfigure}
~
  \begin{subfigure}[b]{0.45\textwidth}
    \includegraphics[width=\textwidth]{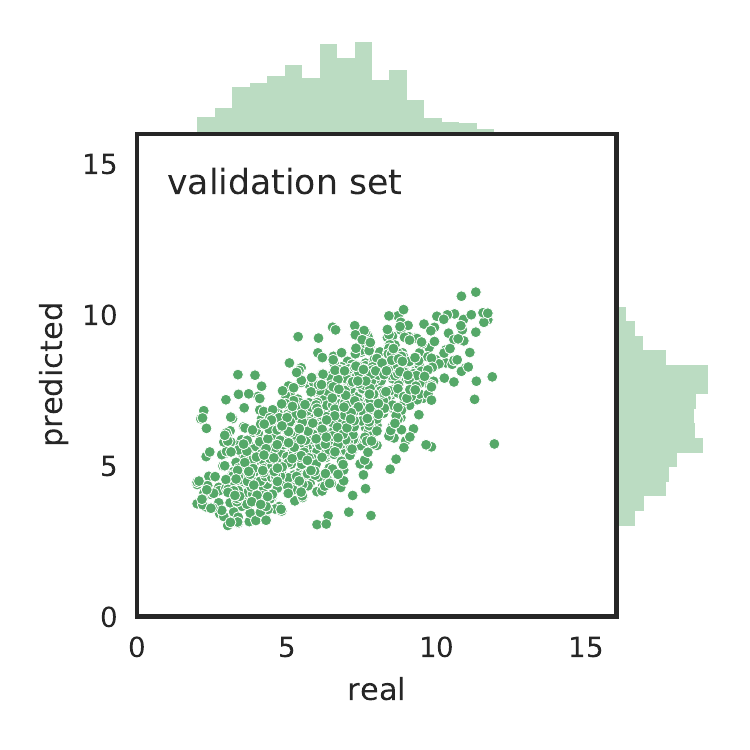}
    \caption{validation set; $RMSE=1.44$, $R=0.72$}
    \label{fig:pred-val}
  \end{subfigure}

\caption{Predictions for two test sets (core sets from PDBbind v. 2016 and v. 2013), training set and validation set.}
\label{fig:pred}
\end{figure*}

After 14 epochs of training, the model started to overfit, and the error on the validation set started to slowly yet steadily increase.
The best set of weights of the network, obtained after 14 epochs of training, was saved and used as the final model.
Model performance was evaluated on all subsets of the data (see Table~\ref{tab:results} and Figure~\ref{fig:pred}).
For each complex in the dataset, affinity was predicted and compared to the real value.
Prediction error was measured with $RMSE$ and $MAE$ (mean absolute error).
The correlation between the scores and experimentally measured binding constants was assessed with the Pearson's correlation coefficient ($R$) and the standard deviation in regression ($SD$).
$SD$ is a measure used in CASF \citep{Li2014} and is defined as follows:
\[
SD = \sqrt[]{\frac{1}{N-1} \sum_{i=1}^{N} \left[ t_i - \left(a y_i + b \right)\right] ^2 }
\]
where $t_i$ and $y_i$ are the measured and predicted affinities for the $i$\textsuperscript{th} complex, whereas $a$ and $b$ are the slope and the intercept of the regression line between measured and predicted values, respectively.

\begin{table}[!ht]
\begin{tabular}{lllll}
\hline
Dataset                        & $RMSE$ & $MAE$ & $SD$  & $R$  \\
\hline
v. 2013 core set               & 1.62   & 1.31  & 1.61  & 0.70 \\
v. 2016 core set               & 1.42   & 1.13  & 1.37  & 0.78 \\
Validation                     & 1.44   & 1.14  & 1.43  & 0.72 \\
Training                       & 1.21   & 0.95  & 1.19  & 0.77 \\
Baseline (X-Score) & -      & -     & 1.78  & 0.61 \\
\hline
\end{tabular}%}{$^\text{a}$ reported in~\citep{Li2014}}
\caption{Pafnucy's performance. Prediction accuracy for each subset was evaluated using four different metrics (see main text).}
\label{tab:results} 
\end{table}

As expected, the network achieves the lowest error on the training set (Figure~\ref{fig:pred-train}), which was used to find the weights of the network.
More importantly, Pafnucy also returns accurate predictions for the two test sets (Figures~\ref{fig:pred-test} and~\ref{fig:pred-test2013}), which were unknown to the model during training and validation.
The results on the v. 2013 core set, although substantially worse than for other subsets, are still better than those for any other scoring function tested by \citet{Li2014} -- the best-performing X-Score had $R=0.61$ and $SD=1.78$, while our model achieved $R=0.70$ and $SD=1.61$ (see Table~\ref{tab:results}).

We also compared Pafnucy to X-Score on the Astex Diverse Set (Table~\ref{tab:astex}).
This experiment provides Pafnucy with a test set completely separate from the data provided by Liu \emph{et~al}.

\begin{table}[!ht]
\begin{tabular}{lllll}
\hline
Method     & $RMSE$ & $MAE$ & $SD$  & $R$  \\
\hline
Pafnucy    & 1.43   & 1.13  & 1.43  & 0.57 \\
X-Score    & 1.55   & 1.22  & 1.48  & 0.52 \\
\hline
\end{tabular}
\caption{Predictions accuracy on the Astex Diverse Set.}
\label{tab:astex}
\end{table}

Both methods have comparable errors to those obtained on the PDBbind data.
As expected, Pafnucy outperforms X-Score on the Astex Diverse Set, regardless of which measure is used.
The observed correlation, however, is lower for both methods.
This effect is partially due to the fact that the Astex dataset contains only 73 complexes, and therefore, correlation is much more sensitive to small changes in the predictions than for bigger subsets.

%%%%%%%%%%%%%%%%%%%%%%%%%%%%%%%%%%%%%%%%%%%%%%%%%%%%%%%%%%%%%%%%%%%%%%%%%%%%%
% DISCUSSION
%%%%%%%%%%%%%%%%%%%%%%%%%%%%%%%%%%%%%%%%%%%%%%%%%%%%%%%%%%%%%%%%%%%%%%%%%%%%%

\section{Discussion}\label{sec:discussion}
\subsection{Stability of the results with respect to input rotation}\label{sec:stability}
One of the biggest challenges of this project was to create a model that was insensitive to the orientation of a molecular complex.
The model presented in this work is not rotation-invariant, similar to the 2D convolutional neural networks used in image recognition; the input looks differently when an object is shown from a different angle, yet it contains the same information about the underlying real object.
Therefore, to generalize well, the model needed to learn to extract this information from differently presented input.
In order to achieve this outcome, we augmented the dataset with systematic rotations of the input data.
If Pafnucy was trained correctly, it should return similar predictions regardless of the orientation of the complex.

To test the model's stability we selected the PDE10A protein, a cAMP/cGMP phosphodiesterase important in signal transduction and recently linked to neuropsychiatric disorders \citep{Macmullen2017}.
PDE10A is complexed with 57 different ligands in the PDBBind database (41 complexes in the training set, 6 in the validation and 10 in the test set).
Each of the complexes was presented to the model in 24 different rotations, and the distribution of returned predictions was analyzed.
As anticipated, the variability of the predicted binding constants is low (see Supplementary Figure~S4).
Additionally, the variability does not depend on the value of the prediction nor on the subset to which the molecule belongs.

\subsection{How Pafnucy sees and processes the data}\label{sec:modified}
Neural networks are often deemed harder to analyze and interpret than simpler models and are sometimes regarded as ``black-boxes''.
The worry is that a model can yield good predictions for the wrong reasons (e.g., artefacts hidden in the data) and therefore will not generalize well for new datasets.
In order to trust a neural network and its predictions, one needs to ensure that the model uses information that is relevant to the task at hand.
In this section, we analyze which parts of the input are the most important and have the biggest impact on the predictions.

\begin{figure}[!htb]
  \centering
  \includegraphics{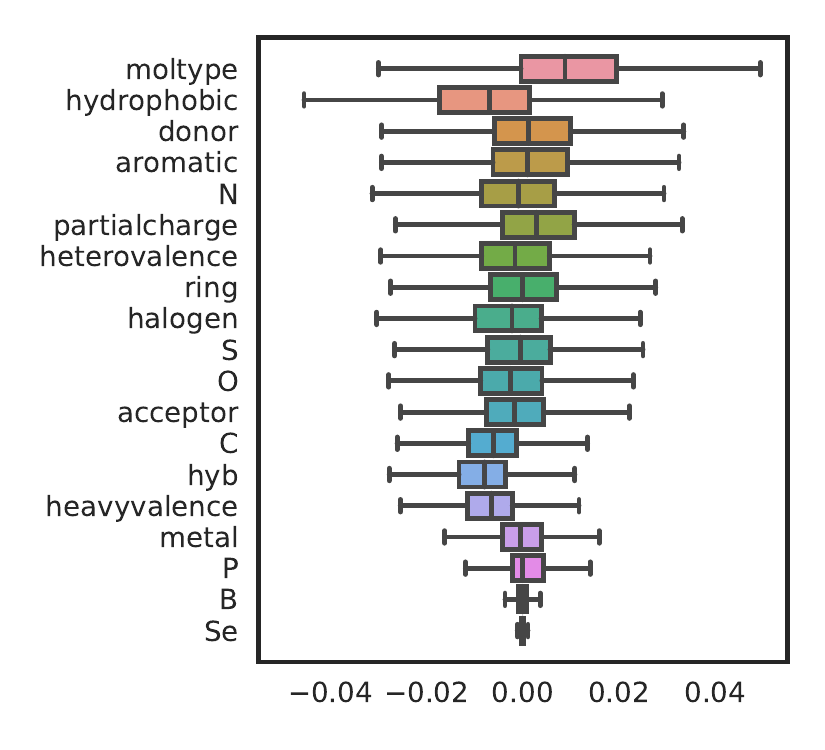}
  \caption[Feature importance]{Range of weights for each input channel (feature). Outliers are not shown.}
  \label{fig:importance}
\end{figure}

In the case of random forests, for example, there is an established way to calculate feature importance based on the impurity decrease \citep{Breiman1984}.
With neural networks, there is no such consensus, as the interpretation of the model's parameters may differ considerably between networks with different architectures.

In the case of Pafnucy, which was trained with L2, we can estimate feature importance by looking at the distributions of weights associated with the convolutional filters in the first hidden layer.
Their initial values were close to 0 (see Section~\ref{sec:train} for more details).
During training, the weights tend to spread and form wider ranges, as weights with higher absolute values pass more information to the deeper layers of the network.
Because Pafnucy was trained with L2 regularization, only crucial weights were likely to have such high absolute values.

The input was represented using 19 channels, some of which were expected to be of low relevance for the model (e.g., the boron atom type).
As we can see in the Figure~\ref{fig:importance}, the feature with the widest range is the \emph{moltype} -- feature distinguishing the protein from the ligand.
This result implies that Pafnucy learned that binding affinity depends on the relationship between the two molecules and that recognizing them is crucial.
Additionally, the weights for selenium and boron atom types (\emph{Se} and \emph{B}, respectively) barely changed during training and are close to 0.
This result can be interpreted in two ways: either the network found other features of protein-ligand complexes more important for binding affinity, or due to infrequent occurrence of these atom types in ligands the network was not able to find any general patterns for their influence on binding affinity.

\begin{figure}[H]
  \centering
  \begin{subfigure}[b]{0.34\textwidth}
    \includegraphics[width=\textwidth]{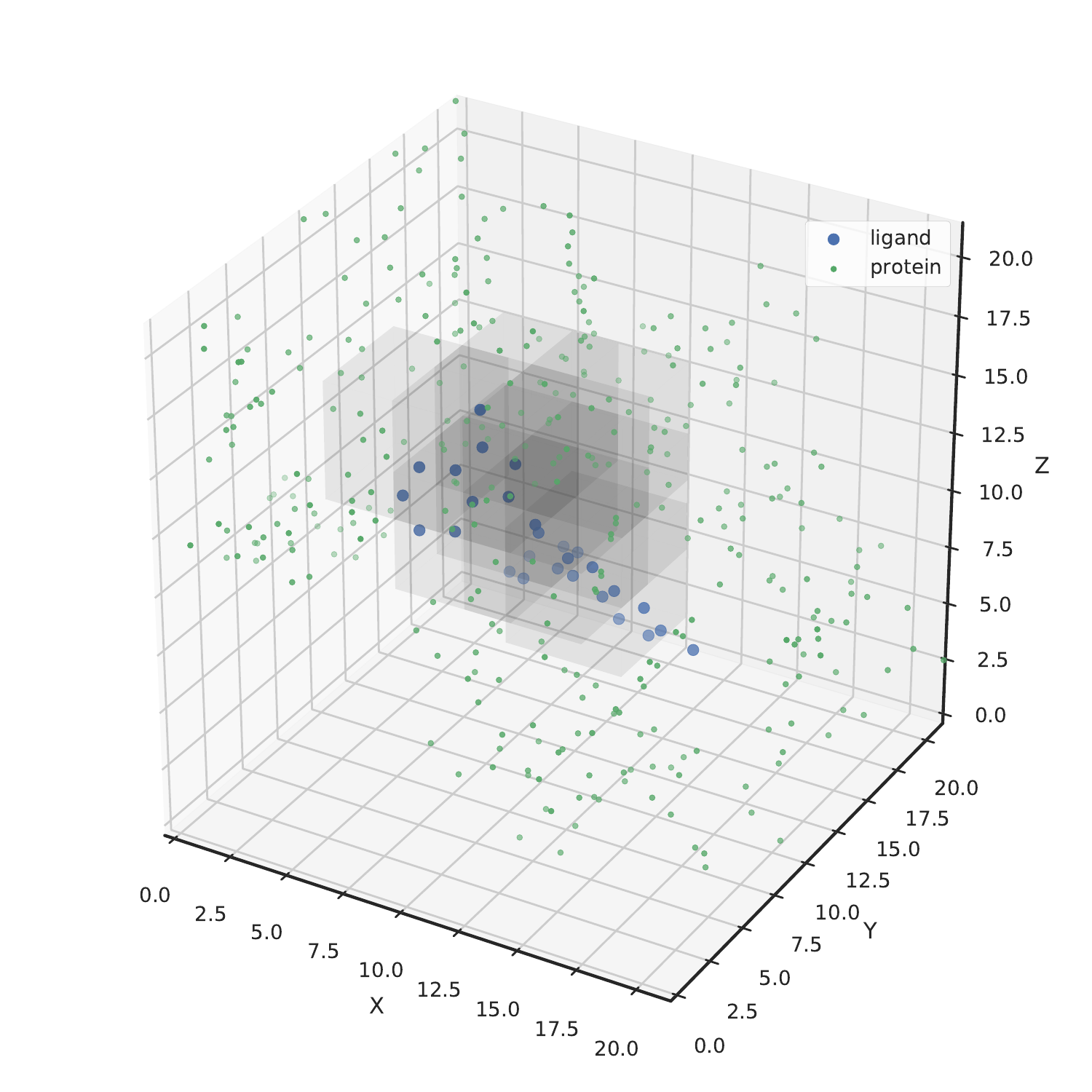}
    \caption{Original orientation}
    \label{fig:change-rot0}
  \end{subfigure}

  \begin{subfigure}[b]{0.34\textwidth}
    \includegraphics[width=\textwidth]{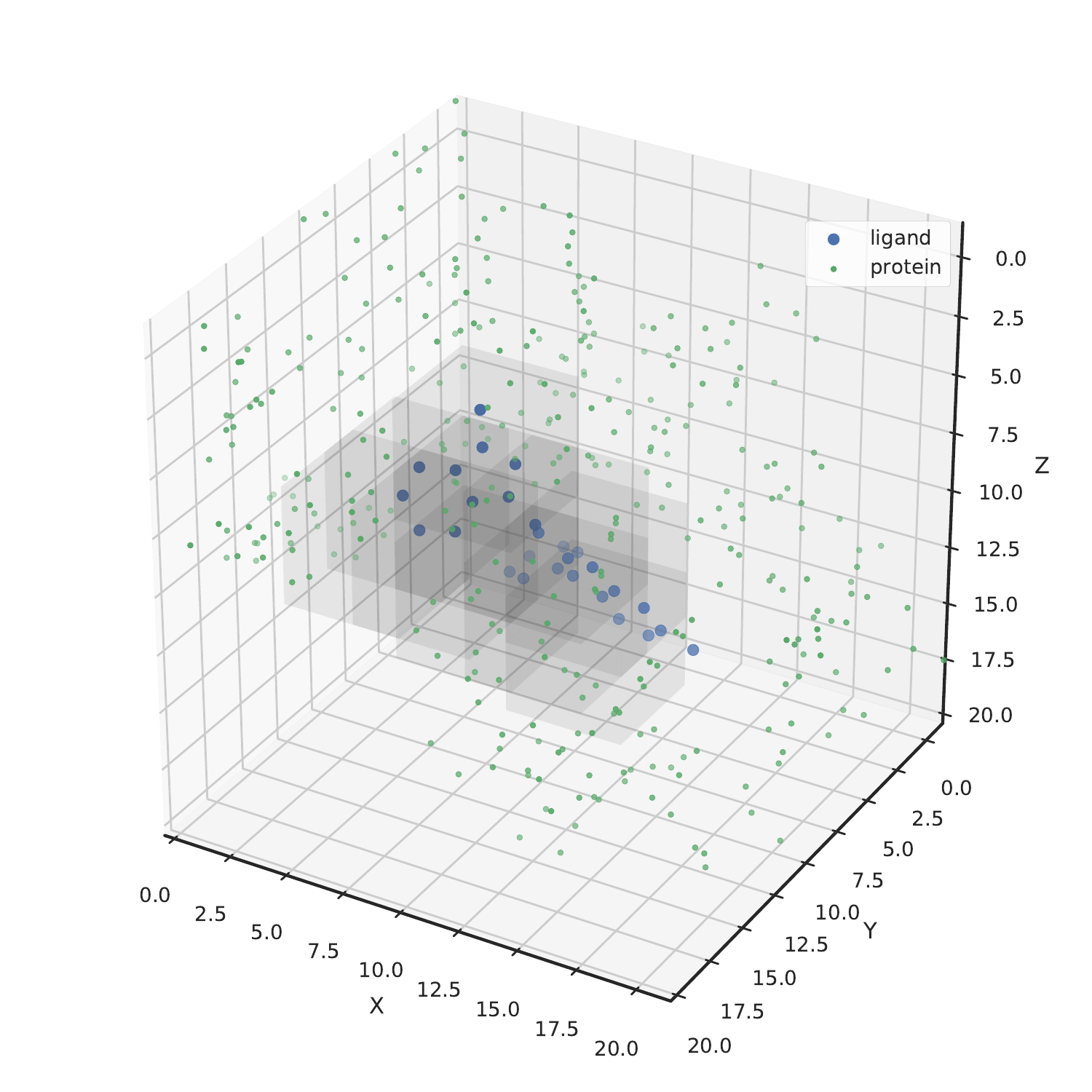}
    \caption{Rotated by $180^{\circ}$ about the X-axis}
    \label{fig:change-rot2}
  \end{subfigure}
  
  \begin{subfigure}[b]{0.34\textwidth}
    \centering
    \includegraphics[width=0.9\textwidth]{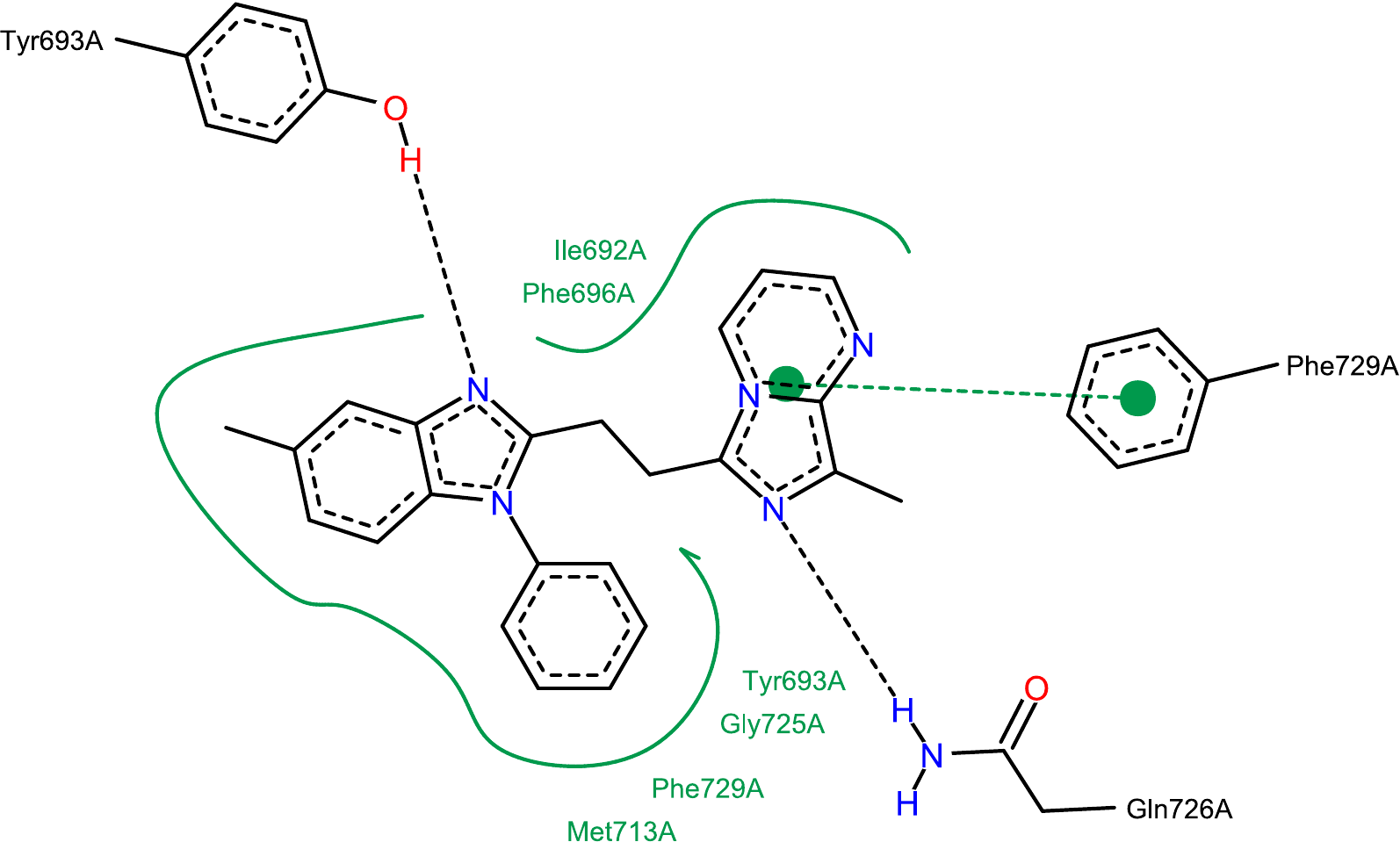}
    \caption{Protein-ligand interactions. Graphic was generated with Poseview \citep{Stierand2010}.}
    \label{fig:change-int}
  \end{subfigure}

\caption{The most important parts of the input. Regardless of the complex orientation, the same region of the input had the highest impact on the prediction. Note that the second plot is rotated back about the X-axis to ease the comparison.}
\label{fig:changes}
\end{figure}

To further inspect how the network utilizes the input, we analyzed the impact of missing data on the prediction.
To inspect this, we selected one of the PDE10A complexes with a benzimidazole inhibitor (complex PDB ID: 3WS8; ligand PDB ID: X4C).
The experiment was carried out as follows: we produced 343 corrupted complexes with some missing data and predicted the binding affinity for each.
The missing data were produced by deleting a 5-\r{A} cubic box from the original data.
We slid the box with a 3-\r{A} step (in every direction), thus yielding $7^3 = 343$ corrupted inputs.
Next, we rotated the complex by $180^\circ$ about the X-axis and followed the same procedure, thus yielding another 343 corrupted inputs.
Then, for each of the two orientations, we took 10 corrupted inputs that had the highest drop in predicted affinity (Figure~\ref{fig:changes}).
We wanted to find which atoms' absence caused the highest drops in the predictions.

% {{\parfillskip0pt\par}}

% \noindent 

As we can see in Figures~\ref{fig:change-rot0} and~\ref{fig:change-rot2}, for both orientations, we identified the same region containing the ligand and its nearest neighbourhood.
The boxes contain the amino-acids participating in the interactions with the ligand, i.e., Gln726, which forms a hydrogen bond, and Phe729, which forms a $\pi-\pi$ interaction with the ligand (Figure~\ref{fig:change-int}).

Additionally, if we considered 15 corrupted complexes with the highest drop in predictions, we find other amino-acids interacting with the ligand: Tyr693, which forms a hydrogen bond, and Met713, which forms hydrophobic contacts with the ligand.
The methodology presented above can be applied to other complexes in order to elucidate specific ligand-receptor interactions with the most profound effect on the prediction.

Going back to the uncorrupted input, we wanted to investigate how Pafnucy managed to give almost identical predictions for two different orientations of the complex (the second rotated about the X-axis by $180^\circ$).
For this inquiry, we analyzed the activations of the hidden layers for the two inputs.

\begin{figure}[!htb]
  \includegraphics[width=0.45\textwidth]{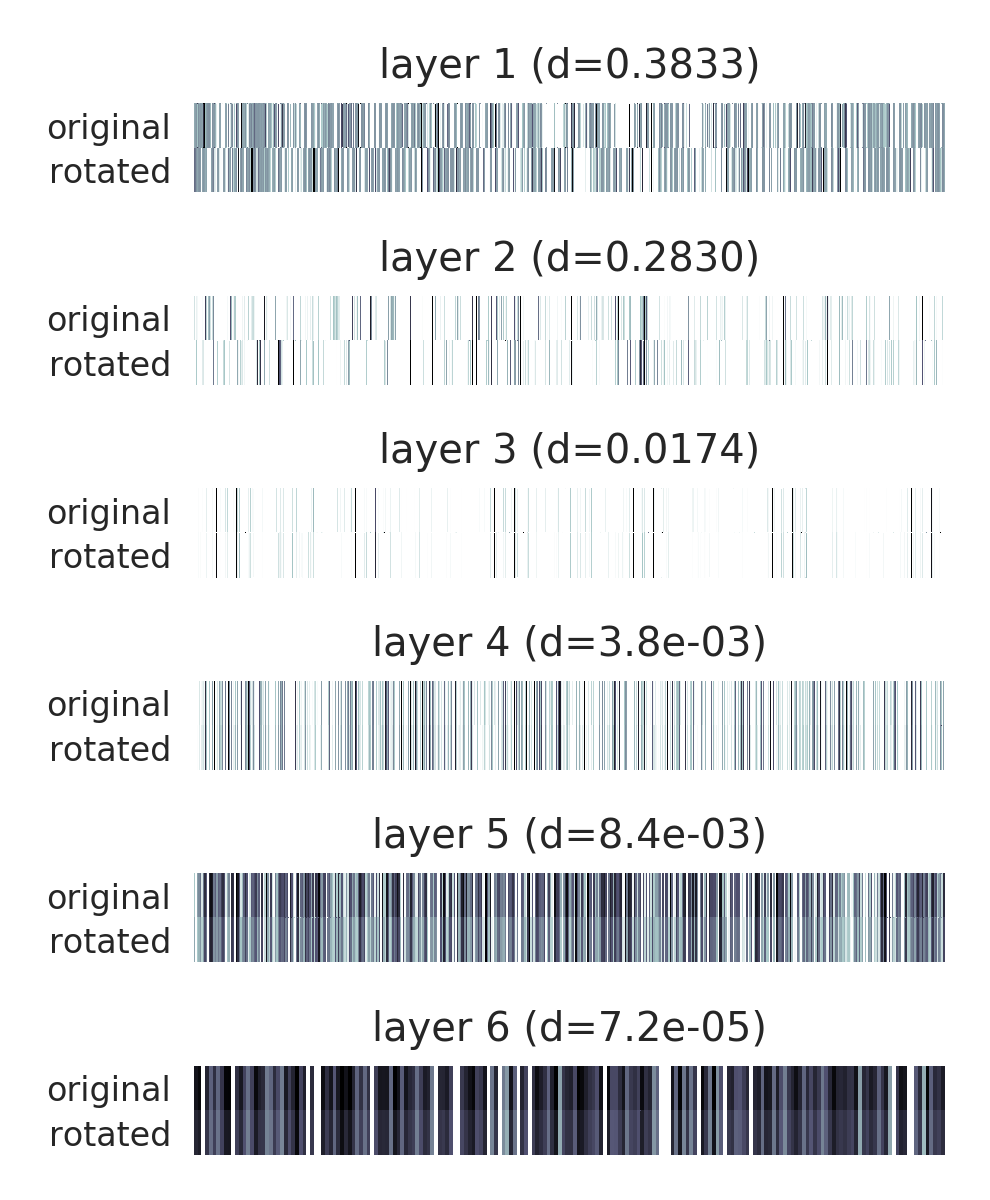}
  \caption[Activations]{Activations on the hidden layers for two orientations of the PDE10A complex (PDB ID: 3WS8). Cosine distances between the activation patterns for each layer are provided.}
  \label{fig:activations}
\end{figure}

In Figure~\ref{fig:activations}, we can see that the first hidden layer has very different activation patterns for the two orientations of the input.
Pafnucy gets very different data and needs to use different filters in the first convolutional layer to process them.
However, the closer we get to the output layer, the more similar the activations become.
We can clearly see that our model learned to extract the same information from differently presented data.

\section{Conclusions}
In this work, we presented a deep neural network, Pafnucy, which can be used in structure based ligand discovery campaigns; as a scoring function in virtual screening or affinity predictor for novel molecules after a complex is generated.
The model was tested on the CASF ``scoring power'' benchmark and outperformed all 20 state-of-the-art scoring functions tested by the CASF authors.
The results obtained and the careful analysis of the network show that Pafnucy makes reliable predictions based on relevant features.

Pafnucy and its source code, together with the Jupyter Notebooks used to prepare the data and analyze the results, are freely available at \href{http://gitlab.com/cheminfIBB/pafnucy}{http://gitlab.com/cheminfIBB/pafnucy}.
Usage examples and scripts are also available to facilitate the most common use-cases: preparing the input data, predicting binding affinity, and training a new network.
We hope that these features will make Pafnucy easily applicable and adaptable by other researchers.
Preparing the environment with all needed dependencies and using the model for the new data can be done with minimum effort:
\begin{minipage}{\textwidth}
{\scriptsize
\lstset{frame=L,breaklines=true,breakatwhitespace=true}
\begin{lstlisting}
git clone https://gitlab.com/cheminfIBB/pafnucy
cd pafnucy
conda env create -f environment_gpu.yml
source activate pafnucy_env
python prepare.py -l ligand.mol2 -p pocket.mol2 \
    -o data.hdf
python predict.py -i data.hdf -o predictions.csv
\end{lstlisting}
}
\end{minipage}

\section*{Acknowledgements}
The authors thank Maciej Wójcikowski for providing the data and X-Score results for the Astex Diverse Set and Maciej Dziubiński for his help in revising the manuscript. \vspace*{-12pt}

\section*{Funding}

This work was supported by the Polish Ministry of Science and Higher Education (POIG.02.03.00-00-003/09-00 and IP2010 037470)

\bibliographystyle{natbib}
\bibliography{references}

\appendix
\renewcommand\thefigure{S\arabic{figure}}
\setcounter{figure}{0}

\cleardoublepage
\section*{Supplementary Figures}

\begin{figure}[!htb]
  \includegraphics[width=0.4\textwidth]{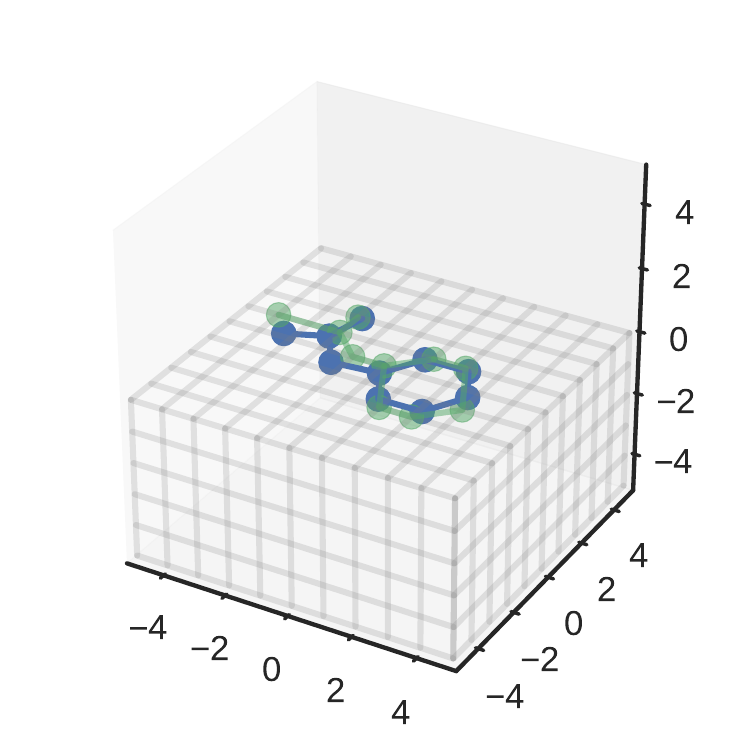}
  \caption[3D grid]{Transforming structure to a 3D grid. Original atom positions are in blue and atom positions in the grid are shown in green. The depicted molecule is relatively flat and Z coordinates are all pulled to a single plane in the grid (Z=0).}
\end{figure}

\begin{figure}[!htb]
  \includegraphics[width=0.45\textwidth]{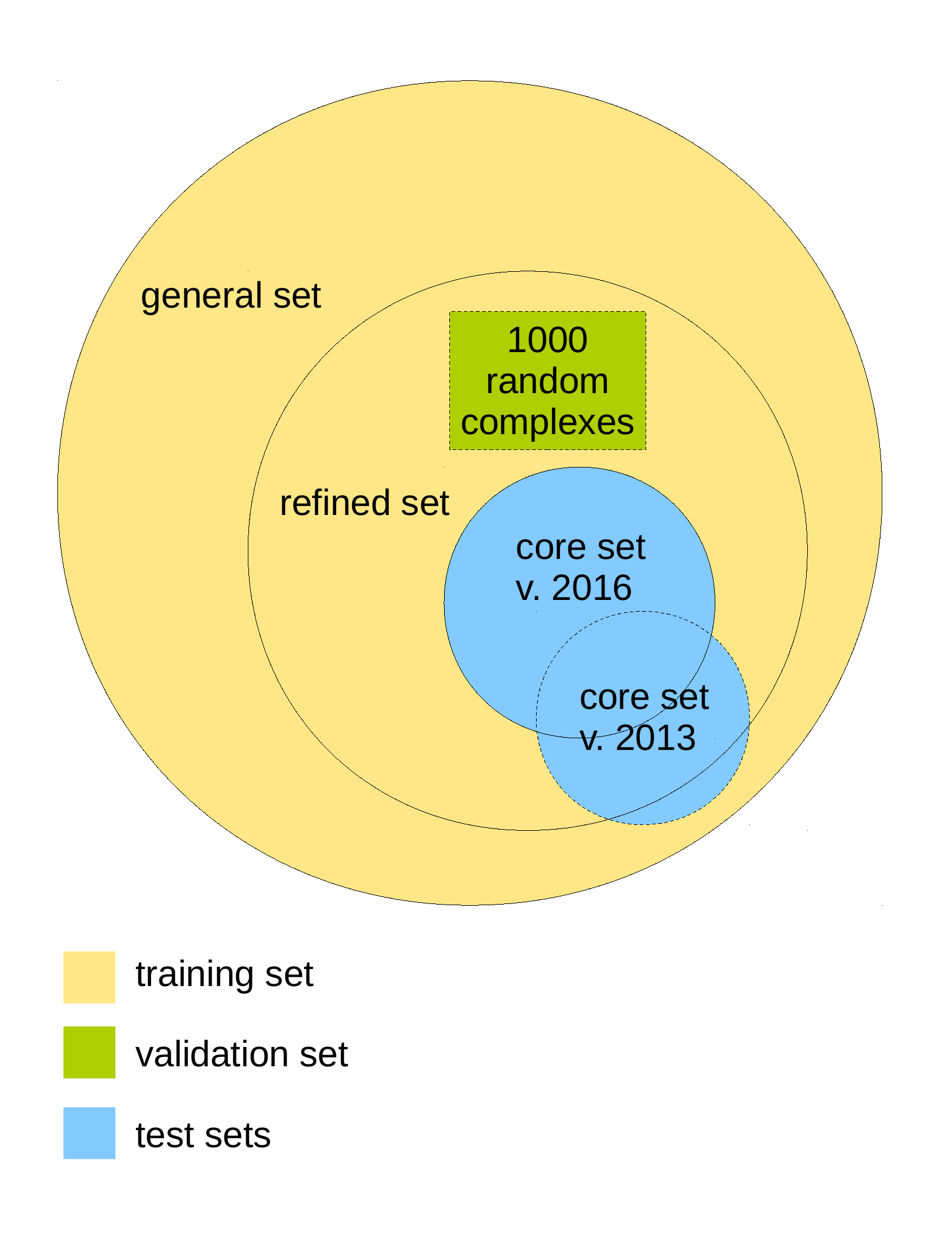}
  \caption[PDBbind]{Scheme illustrating the way the PDBbind database was divided into training, validation, and test set.}
\end{figure}

\begin{figure}[!htb]
  \includegraphics[width=0.47\textwidth]{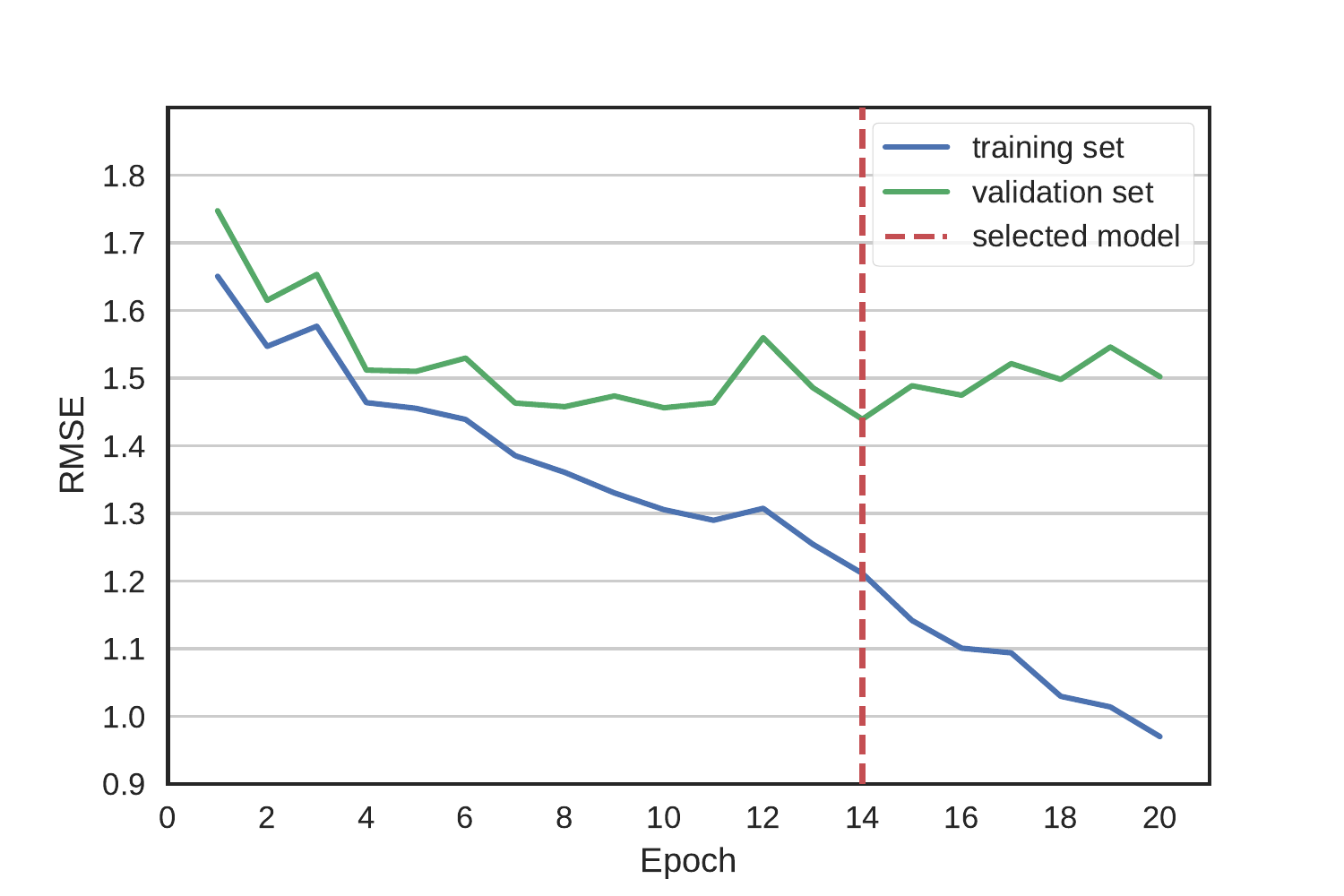}
  \caption[Training curves]{Values of RMSE during training. The error of the untrained model is not shown and it was equal to 4.10 for the training set and 4.26 for the validation set, respectively. Model trained for 14 epochs was selected as the final model based on validation error.}
\end{figure}

\begin{figure}[!htb]
  \includegraphics[width=0.4\textwidth]{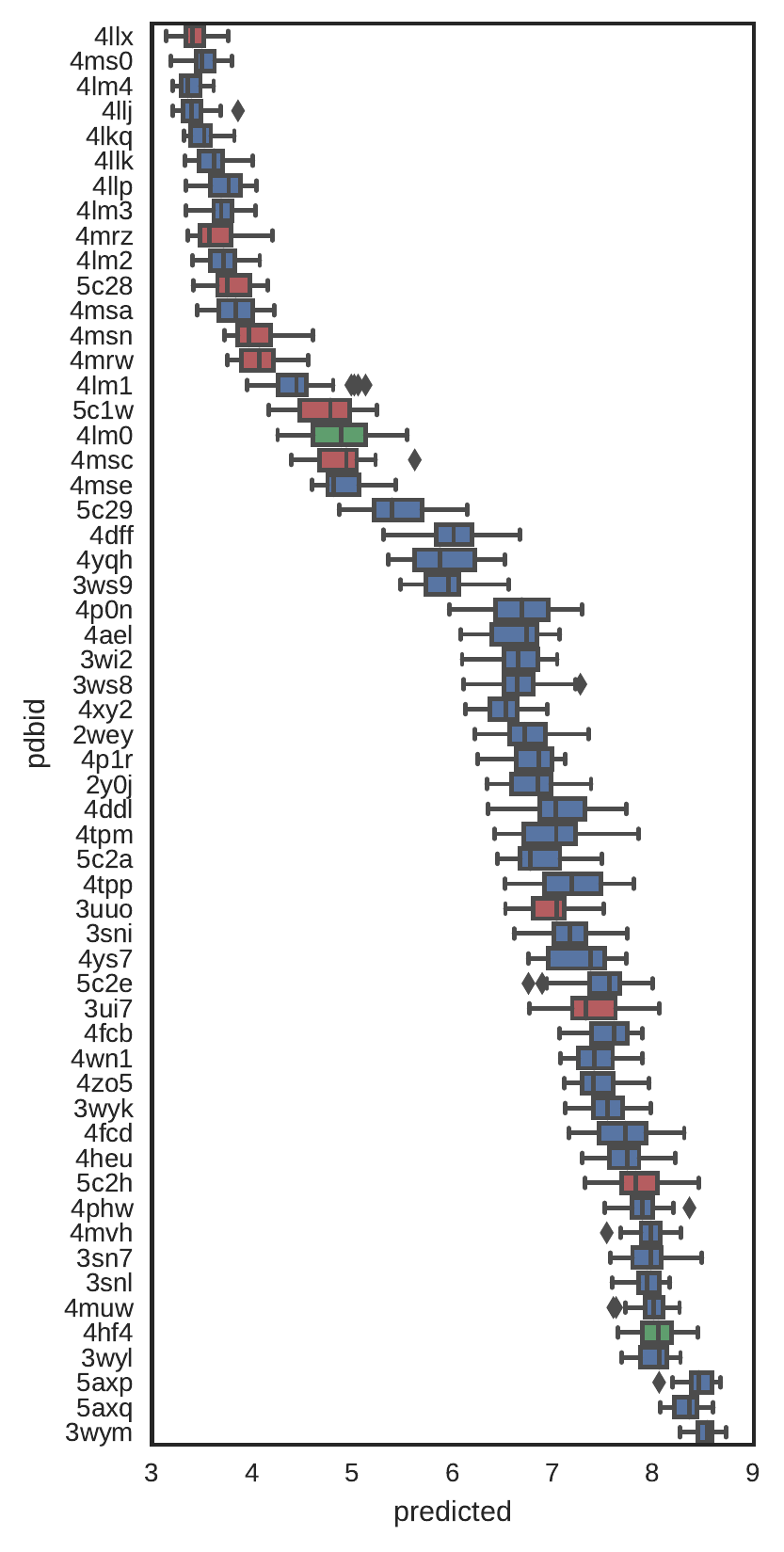}
  \caption[Robustness to rotations]{Stability of the predictions with respect to the rotation of the complex.}
\end{figure}

\end{document}